\documentclass{article} 
\usepackage{iclr2016_workshop,times}
\usepackage{url}
\usepackage{graphicx, subfigure}
\usepackage{amsmath,amsfonts,amssymb,bm}
\usepackage{wrapfig}
\usepackage[hidelinks]{hyperref}
\usepackage{natbib}

\renewcommand{\vec}[1]{\boldsymbol{#1}}

\newcommand{\mat}[1]{\mathbf{#1}}

\usepackage{adjustbox}
\usepackage{array}
\usepackage{booktabs}

\newcolumntype{R}[2]{%
    >{\adjustbox{angle=#1,lap=\width-(#2)}\bgroup}%
    l%
    <{\egroup}%
}

\newcommand{\wv}[1]{\vec{x}_{#1}} 
\newcommand{\hidden}[1]{\vec{h}_{#1}} 
\newcommand{\out}[1]{\vec{y}_{#1}} 
\newcommand{\cont}[1]{\vec{c}_{#1}}
\newcommand{\softmax}[1]{\text{softmax}\left(#1\right)}
\newcommand{\wmat}[1]{\mat{W}_{#1}}
\newcommand{\wvec}[1]{\vec{w}_{#1}}
\newcommand{\bias}{\vec{b}}
\newcommand{\func}[2]{\vec{g}_{#1}\left(#2\right)}
\newcommand{\gfunc}[1]{\vec{s}\left(#1\right)}
\newcommand{\scmh}{{\sc ccDCLM}}
\newcommand{\scmo}{{\sc coDCLM}}
\newcommand{\drnnlm}{{\sc dRnnlm}}
\newcommand{\hrnnlm}{{\sc hRnnlm}}
\newcommand{\rnnlm}{{\sc Rnnlm}}
\newcommand{\dam}{{\sc aDCLM}}
\newcommand{\btheta}{\bm{\theta}}

\newcommand{\newtext}[1]{#1}

\title{Document Context Language Models}

\author{
Yangfeng Ji$^1$, Trevor Cohn$^2$, Lingpeng Kong$^3$, Chris Dyer$^3$ \& Jacob Eisenstein$^1$\\
$^1$ School of Interactive Computing, Georgia Institute of Technology\\
$^2$ Department of Computing and Information Systems, University of Melbourne\\
$^3$ School of Computer Science, Carnegie Mellon University
}

\begin{document}

\maketitle

\begin{abstract}
Text documents are structured on multiple levels of detail: individual words are related by syntax, and larger units of text are related by discourse structure. Existing language models generally fail to account for discourse structure, but it is crucial if we are to have language models that reward coherence and generate coherent texts. We present and empirically evaluate a set of multi-level recurrent neural network language models, called Document-Context Language Models (DCLMs), which incorporate contextual information both within and beyond the sentence. In comparison with sentence-level recurrent neural network language models, the DCLMs obtain slightly better predictive likelihoods, and considerably better assessments of document coherence.
\end{abstract}

\section{Introduction}
\label{sec:intro}

Statistical language models are essential components of natural language processing systems, such as machine translation~\citep{koehn2009statistical}, automatic speech recognition~\citep{jurafsky2000speech}, text generation~\citep{sordoni2015neural} and information retrieval~\citep{manning2008introduction}. Language models estimate the probability of a word for a given context. In conventional language models, context is represented by $n$-grams, so these models condition on a fixed number of preceding words. Recurrent Neural Network Language Models~\citep[RNNLMs;][]{mikolov2010recurrent} use a dense vector representation to summarize context across all preceding words within the same sentence. But context operates on multiple levels of detail: on the syntactic level, a word's immediate neighbors are most predictive; but on the level of discourse and topic, all words in the document lend contextual information.


Recent research has developed a variety of ways to incorporate document-level contextual information. For example, both \citet{mikolov2012context} and \citet{le2014distributed} use topic information extracted from the entire document to help predict words in each sentence; \citet{lin2015hierarchical} propose to construct contextual information by predicting the bag-of-words representation of the previous sentence with a separate model; \citet{wang2015larger} build a bag-of-words context from the previous sentence and integrate it into the Long Short-Term Memory (LSTM) generating the current sentence. These models are all \emph{hybrid} architectures in that they are recurrent at the sentence level, but use a different architecture to summarize the context outside the sentence. 

\begin{figure}
  \centering
  \includegraphics[width=8.0cm]{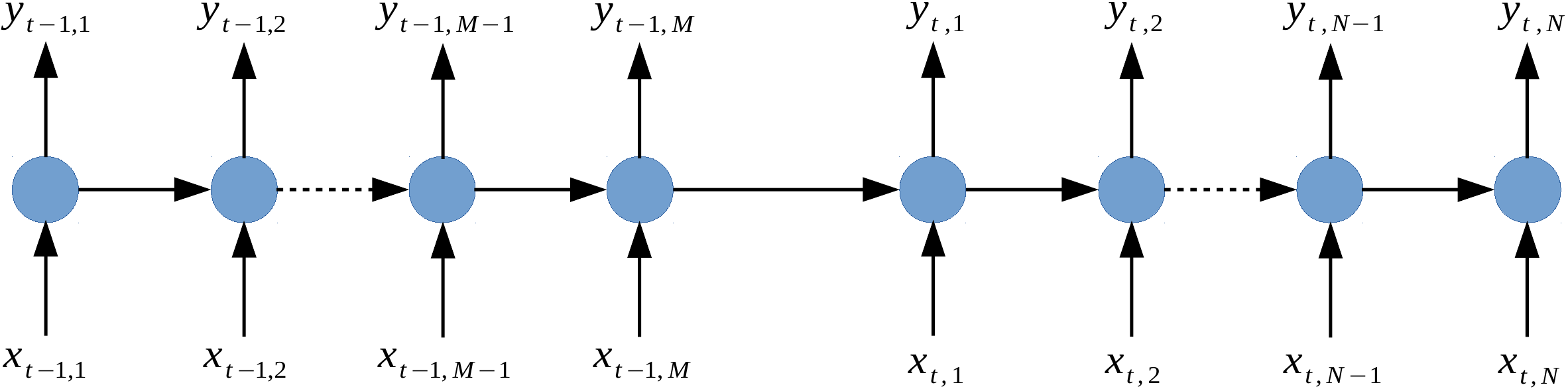}  
  \caption{A fragment of document-level recurrent neural network language model (\drnnlm). It is also an extension of sentence-level RNNLM to the document level by ignoring sentence boundaries.}
  \label{fig:drnnlm}
\end{figure}

In this paper, we explore multi-level recurrent architectures for combining local and global information in language modeling. The simplest such model would be to train a single RNN, ignoring sentence boundaries: as shown in \autoref{fig:drnnlm}, the last hidden state from the previous sentence $t-1$ is used to initialize the first hidden state in sentence $t$. In such an architecture, the length of the RNN is equal to the number of tokens in the document; in typical genres such as news texts, this means training RNNs from sequences of several hundred tokens, which introduces two problems: 
\begin{description}
\item[Information decay] In a sentence with thirty tokens (not unusual in news text), the contextual information from the previous sentence must be propagated through the recurrent dynamics thirty times before it can reach the last token of the current sentence. Meaningful document-level information is unlikely to survive such a long pipeline.
\item[Learning] It is notoriously difficult to train recurrent architectures that involve many time steps~\citep{bengio1994learning}.
In the case of an RNN trained on an entire document, backpropagation would have to run over hundreds of steps, posing severe numerical challenges.
\end{description}
In this paper, we use multi-level recurrent structures to solve both of these problems, thereby successfully efficiently leveraging document-level context in language modeling. We present several variant Document-Context Language Models (DCLMs), and evaluate them on predictive likelihood and their ability to capture document coherence.

\section{Modeling framework}
\label{sec:models}

The core modeling idea of this work is to integrate contextual information from the RNN language model of the previous sentence into the language model of the current sentence. We present three alternative models, each with various practical or theoretical merits, and then evaluate them in \autoref{sec:exp}.

\subsection{Recurrent Neural Network Language Models}
\label{subsec:rnnlm}

We start from a recurrent neural network language model (\rnnlm) to explain some necessary terms. Given a sentence $\{\wv{n}\}_{n=1}^{N}$, a recurrent neural network language model is defined as 
\begin{align}
  \label{eq:rnnlm}
  \hidden{n} =& \func{}{\hidden{n-1},\wv{n}}\\
  \out{n} =& \softmax{\wmat{o}\hidden{n} + \bias},
             \label{eq:rnnlm-emit}
\end{align}
where $\wv{n}\in\mathbb{R}^{K}$ is the distributed representation of the $n$-th word, 
$\hidden{n}\in\mathbb{R}^{H}$ is the corresponding hidden state computed from the word representation and the previous hidden state $\hidden{n-1}$, 
and $\bias$ is the bias term. 
$K$ and $H$ are the input and hidden dimension respectively.
As in the original RNNLM~\citep{mikolov2010recurrent}, $\out{n}$ is a prediction of the $(n+1)$-th word in the sequence.

The transition function $\func{}{\cdot}$ could be any nonlinear function used in neural networks, such as the elementwise sigmoid function, or more complex recurrent functions such as the LSTM~\citep{hochreiter1997long} or GRU~\citep{chung2014empirical}. 
In this work, we use LSTM, as it consistently gives the best performance in our experiments. 
By stacking two LSTM together, we are able obtain a even more powerful transition function, called multi-layer LSTM~\citep{sutskever2014sequence}. 
In a multi-layer LSTM, the hidden state from a lower-layer LSTM cell is used as the input to the upper-layer, and the hidden state from the final-layer is used for prediction. 
In our following models, we fix the number of layers as two.

In the rest of this section, we will consider different ways to employ the contextual information for document-level language modeling. All models obtain the contextual representation from the hidden states of the previous sentence, but they use this information in different ways.

\subsection{Model I: Context-to-Context DCLM}
\label{subsec:ccdclm}

\begin{figure}
  \centering
  \subfigure[\scmh]{
    \includegraphics[width=6.5cm]{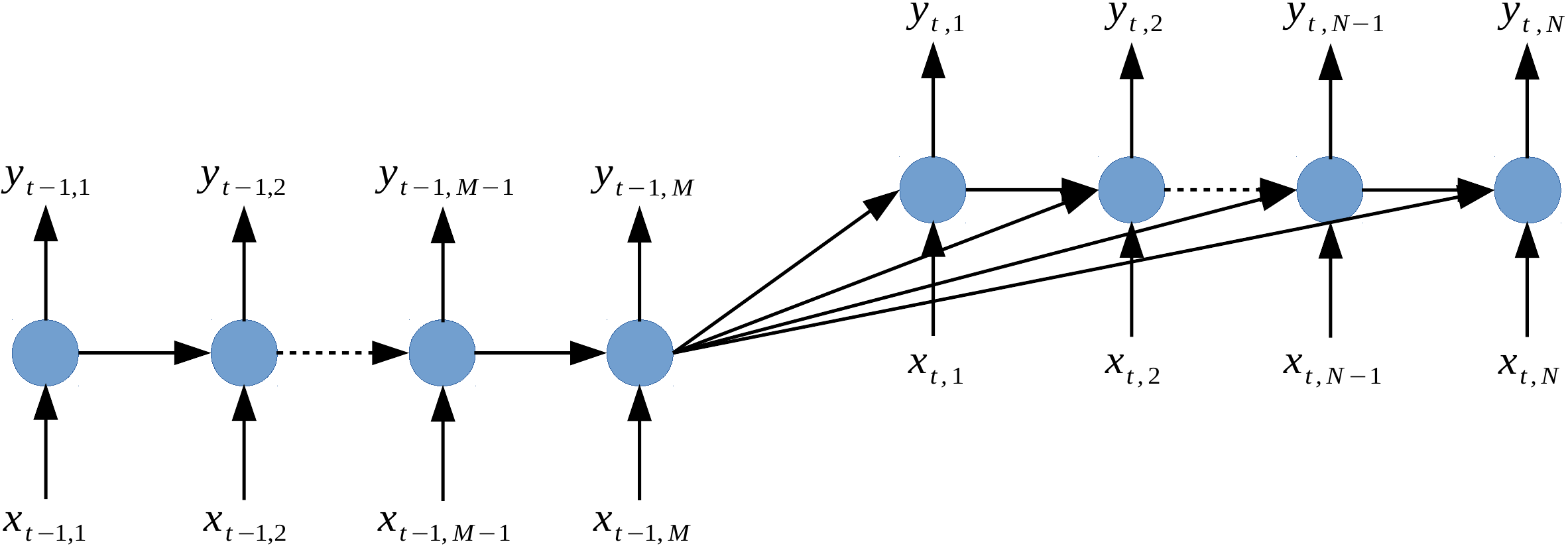}
    \label{fig:scm-h}
  }
  \quad
  \subfigure[\scmo]{
    \includegraphics[width=6.5cm]{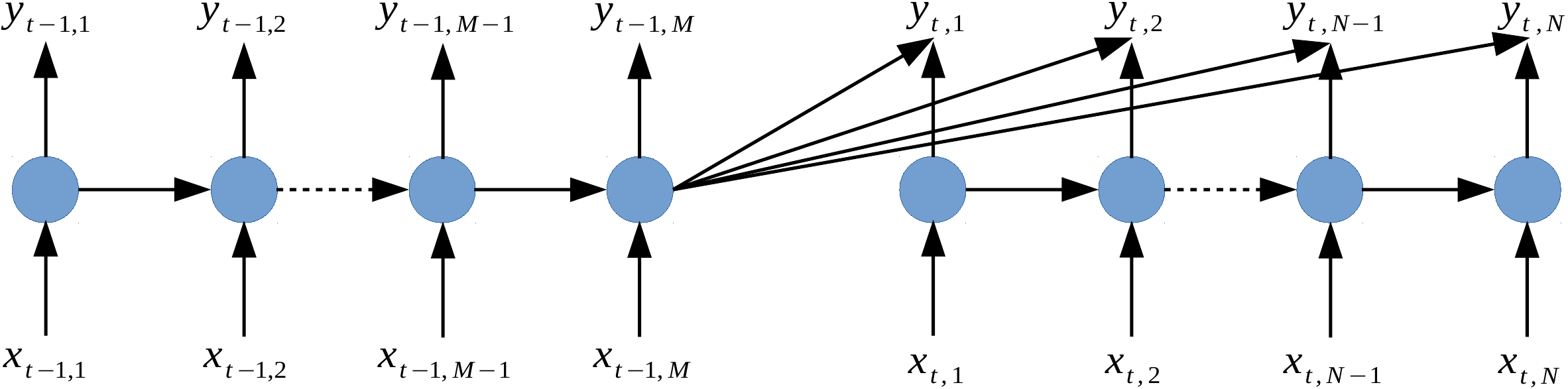}
    \label{fig:scm-o}
  }
  \caption{Context-to-context and context-to-output DCLMs}
  \label{fig:model}
\end{figure}

The underlying assumption of this work is that contextual information from previous sentences needs to be able to ``short-circuit'' the standard RNN, so as to more directly impact the generation of words across longer spans of text. We first consider the relevant contextual information to be the final hidden representation from the previous sentence $t-1$, so that,
\begin{equation}
\cont{t-1} = \hidden{t-1,M}
\label{eq:cont-t-t1}
\end{equation}
where $M$ is the length of sentence $t-1$. We then create additional paths for this information to impact each hidden representation in the current sentence $t$. Writing $\wv{t,n}$ for the word representation of the $n$-th word in the $t$-th sentence, we have,
\begin{align}
\label{eq:hidden}
\hidden{t,n} = &  \func{\btheta}{\hidden{t,n-1},\gfunc{\wv{t,n},\cont{t-1}}}
\end{align}
where $\func{\btheta}{\cdot}$ is the activation function parameterized by $\theta$ and $\gfunc{\cdot}$ is a function that combines the context vector with the input $\wv{t,n}$ for the hidden state. In future work we may consider a variety of forms for this function, but here we simply concatenate the representations,
\begin{equation}
\gfunc{\wv{t,n},\cont{t-1}} = [\wv{t,n},\cont{t-1}].
\end{equation}
The emission probability for $\out{t,n}$ is then computed from $\hidden{t,n}$ as in the standard RNNLM (\autoref{eq:rnnlm-emit}). The underlying assumption of this model is that contextual information should impact the generation of each word in the current sentence.  The model therefore introduces computational ``short-circuits'' for cross-sentence information, as illustrated in Figure~\ref{fig:scm-h}.
Because information flows from one hidden vector to another, we call this the \textbf{context-to-context Document Context Language Model}, abbreviated \scmh.

With this specific architecture, the number of parameters is $H(16H+3K+6) + V(H+K+1)$, where $H$ is the size of the hidden representation, $K$ is the size of the word representation, and $V$ is the vocabulary size. The constant factors come with the weight matrices within a two-layer LSTM unit.
This is in the same complexity class as the standard RNNLM.
Special handling is necessary for the first sentence of the document. 
Inspired by the idea of sentence-level language modeling, we introduce a \emph{dummy} contextual representation $\cont{0}$ as a {\sc Start} symbol for a document. This is another parameter to be learned jointly with the other parameters in this model.


The training procedure of \scmh~is similar to a conventional RNNLM: we move from left to right through the document and compute a softmax loss on each output $\out{t,n}$. We then backpropagate this loss through the entire sequences.


\subsubsection{Model II: Context-to-Output DCLM}
Rather than incorporating the document context into the recurrent definition of the hidden state, we can push it directly to the output, as illustrated in Figure~\ref{fig:scm-o}. Let $\hidden{t,n}$ be the hidden state from a conventional RNNLM of sentence $t$, 
\begin{equation}
  \hidden{t,n} = \func{\theta}{\hidden{t,n-1}, \wv{t,n}}.
\end{equation}
Then, the context vector $\cont{t-1}$ is directly used in the output layer as 
\begin{equation}
  \label{eq:out-tn-o}
  \out{t,n}\sim\softmax{\wmat{h}\hidden{t,n} + \wmat{c}\cont{t-1} + \bias}
\end{equation}
where $\cont{t-1}$ is defined in \autoref{eq:cont-t-t1}. Because the document context impacts the output directly, we call this model the {\bf context-to-output DCLM} (\scmo).
The modification on the model architecture from \scmh~to \scmo~leads to a notable change on the number of parameters. 
The total number of parameters of \scmo~is $H(13H+3K+6) + V(2H+K+1)$. 
The difference of the parameter numbers between these \scmo~and \scmh~is $VH-3H^2$. Recall that $V$ is the vocabulary size and $H$ is the size of latent representation, in most cases we have $V \geq 10^4$ and $H \approx 10^2$. Therefore $V \gg H$ in all reasonable cases, and \scmo~includes more parameters than \scmh~in general.

While the \scmo\ has more parameters that must be learned, it has a potentially important computational advantage. By shifting $\cont{t-1}$ from hidden layer to output layer, the relationship of any two hidden vectors $\hidden{t}$ and $\hidden{t'}$ from different sentences is decoupled, so that each can be computed in isolation. In a guided language generation scenario such as machine translation or speech recognition --- the most common use case of neural language models --- this means that decoding decisions are only \emph{pairwise dependent} across sentences. This is in contrast with the \scmh, where the tying between each $\hidden{t}$ and $\hidden{t+1}$ means that decoding decisions are \emph{jointly dependent} across the entire document. 
This joint dependence may have important advantages, as it propagates contextual information further across the document; the \scmh\ and \scmo\  thereby offer two points on a tradeoff between accuracy and decoding complexity.

\subsection{Attentional DCLM}
One potential shortcoming of \scmh~and \scmo~is the limited capacity of the context vector, $\cont{t-1}$, which is a fixed dimensional representation of the context. While this might suffice for short sentences, as sentences grow longer, the amount of information needing to be carried forward will also grow, and therefore a fixed size embedding may be insufficient. For this reason, we now consider an \emph{attentional} mechanism, based on conditional language models for translation \citep{sutskever2014sequence,bahdanau2015neural} which allows for a dynamic capacity representation of the context.

Central to the attentional mechanism is the context representation, which is defined separately for each word position in the output sentence,
\begin{eqnarray}
  \label{eq:cont-attentional}
  \cont{t-1,n} &=& \sum_{m=1}^M \alpha_{n,m} \hidden{t-1,m} \\
  \vec{\alpha}_n &=& \softmax{ \vec{a}_n } \\
  a_{n,m} &=& \wvec{a}^\top \tanh\left( \wmat{a1} \hidden{t,n} + \wmat{a2} \hidden{t-1,m} \right)
\end{eqnarray}
where $\cont{t-1,n}$ is formulated as a weighted linear combination of all the hidden states in the previous sentence, with weights $\alpha$ constrained to lie on the simplex using the softmax transformation.
Each weight $\alpha_{n,m}$ encodes the importance of the context at position $m$ for generating the current word at $n$,  defined as a neural network with a hidden layer and a single scalar output.
Consequently each position in the generated output can `attend' to different elements of the context sentence, which would arguably be useful to shift the focus to make best use of the context vector during generation. 


The revised definition of the context in \autoref{eq:cont-attentional} requires some minor changes in the generating components. We  include this as an additional input to both the recurrent function (similar to \scmh), and output generating function (akin to \scmo), as follows
\begin{eqnarray}
  \hidden{t,n} &=& \func{\btheta}{\hidden{t,n-1}, \left[\cont{t-1,n}^\top, \wv{t,n}^\top\right]^\top} \\
  \out{t,n} &\sim& \softmax{\wmat{o} \tanh\left( \wmat{h} \hidden{t,n} + \wmat{c} \cont{t-1,n} + \bias{} \right) }
\end{eqnarray}
where the output uses a single hidden layer network to merge the local state and context, before expanding the dimensionality to the size of the output vocabulary, using $\wmat{o}$. The extended model is named as {\bf attentional DCLM} (\dam).




\section{Data and Implementation}
\label{sec:data}
We evaluate our models with perplexity and document-level coherence assessment. The first data set used for evaluation is the Penn Treebank (PTB) corpus~\citep{marcus1993building}, which is a standard data set used for evaluating language models~\citep[e.g.,][]{mikolov2010recurrent}. 
We use the standard split: sections 0-20 for training, 21-22 for development, and 23-24 for test~\citep{mikolov2010recurrent}. We keep the top 10,000 words to construct the vocabulary, and replace lower frequency words with the special token {\sc Unknown}. The vocabulary also includes two special tokens {\sc Start} and {\sc End} to indicate the beginning and end of a sentence. In total, the vocabulary size is $10,003$. 

To investigate the capacity of modeling documents with larger context, we use a subset of the North American News Text (NANT) corpus~\citep{mcclosky2008bllip} to construct another evaluation data set. As shown in~\autoref{tab:dataset}, the average length of the training documents is more than 30 sentences. We follow the same procedure to preprocess the dataset as for the PTB corpus, and keep the top 15,000 words from the training set in the vocabulary. Some basic statistics of both data sets are listed in \autoref{tab:dataset}. 

\begin{table}
  \centering
  {\footnotesize
  \begin{tabular}{lllll}
    \toprule
    & & & \multicolumn{2}{l}{Average Document Length}\\
    \cmidrule{4-5}
    & & \# Documents & \# Tokens & \# Sentences\\
    \midrule
    PTB & Training & 2,000 & 502 & 21 \\
    & Development & 155 & 516 & 22\\
    & Test & 155 & 577 & 24\\
    \midrule
    NANT & Training & 26,462 & 783 & 32 \\
    & Development & 148 & 799 & 33 \\
    & Test & 2,753 & 778 & 32 \\
    \bottomrule
  \end{tabular}
  }
  \caption{Basic statistics of the Penn Treebank (PTB) and North American News Text (NANT) data sets}
  \label{tab:dataset}
\end{table} 
 
\subsection{Implementation}
\label{subsec:imp}

We use a two-layer LSTM to build the recurrent architecture of our document language models, which we implement in the \textsc{cnn} package (\url{https://github.com/clab/cnn}). 
The rest of this section includes some additional details of our implementation, which is available online at \url{https://github.com/jiyfeng/dclm}.

{\bf Initialization} All parameters are initialized with random values drawn from the range $[-\sqrt{6/(d_1+d_2)},\sqrt{6/(d_1+d_2)}]$, where $d_1$ and $d_2$ are the input and output dimensions of the parameter matrix respectively, as suggested by~\cite{glorot2010understanding}.

{\bf Learning} Online learning was performed using AdaGrad~\citep{duchi2011adaptive} with the initial learning $\lambda=0.1$. To avoid the exploding gradient problem, we used the norm clipping trick proposed by \citet{pascanu2012difficulty} and fixed the norm threshold as $\tau=5.0$.

{\bf Hyper-parameters} Our models include two tunable hyper-parameters: the dimension of word representation $K$ and the hidden dimension of LSTM unit $H$. We consider the values $\{32, 48, 64, 96, 128, 256\}$ for both $K$ and $H$. The best combination of $K$ and $H$ for each model is selected by the development sets via grid search. In all experiments, we fix the hidden dimension of the attentional component in \dam\ as 48.

{\bf Document length} As shown in \autoref{tab:dataset}, the average length of documents is more than 500 tokens, with extreme cases having over 1,000 tokens. In practice, we noticed that training on long documents leads to a very slow convergence. We therefore segment documents into several non-overlapping shorter documents, each with at most $L$ sentences, while preserving the original sentence order.
The value of $L$ used in most experiments is $5$, although we compare
with $L=10$ in \autoref{subsec:ppl}.



\section{Experiments}
\label{sec:exp}
We compare the three DCLM-style models (\scmh, \scmo, \dam) with the following competitive alternatives:

\begin{description}
\item[Recurrent neural network language model (\rnnlm)] The model is trained on individual sentences without any contextual information~\citep{mikolov2010recurrent}. The comparison between our models and this baseline system highlights the contribution of contextual information. 
\item[RNNLM w/o sentence boundary (\drnnlm)] This is a straightforward extension of sentence-level RNNLM to document-level, as illustrated in~\autoref{fig:drnnlm}. It can also be viewed a conventional RNNLM without considering sentence boundaries. The difference between \rnnlm~and \drnnlm~is that \drnnlm~is able to consider (a limited amount of) extra-sentential context.
\item[Hierarchical RNNLM (\hrnnlm)] We also adopt the model architecture of HRNNLM~\citep{lin2015hierarchical} as another baseline system, and reimplemented it with several modifications for a fair comparison. Comparing to the original implementation~\citep{lin2015hierarchical}, we first replace the sigmoid recurrence function with a long short-term memory (LSTM) as used in DCLMs. Furthermore, instead of using pretrained word embedding, we update word representation during training. Finally, we jointly train the language models on both sentence-level and document-level. These changes resulted in substantial improvements over the original version of the \hrnnlm; they allow us to isolate what we view as the most substantive difference between the DCLM and this modeling approach, which is how contextual information is identified and exploited.
\end{description}

\subsection{Perplexity}
\label{subsec:ppl}
\newtext{To make a fair comparison across different models, we follow the conventional way to compute perplexity. Particularly, the {\sc Start} and {\sc End} tokens are only used for notational convenience. The {\sc End} token from the previous sentence was never used to predict the {\sc Start} token in the current sentence. Therefore, we have the same computation procedure on perplexity for the models with and without contextual information.
}

\begin{table}
  \centering
  {\footnotesize
    \begin{tabular}{llllll}
      \toprule
      & \multicolumn{2}{l}{PTB} & & \multicolumn{2}{l}{NANT}\\
      \cmidrule{2-3} \cmidrule{5-6}
      Model & Dev & Test & & Dev & Test\\
      \midrule
      {\em Baselines}\\
      1. RNNLM~\citep{mikolov2010recurrent} & 69.24 & 71.88 & & 109.48 & 194.43\\
      2. RNNLM w/o sentence boundary (\drnnlm) & 65.27 & 69.37 & & 101.42 & 181.62 \\
      3. Hierarchical RNNLM (\hrnnlm)~\citep{lin2015hierarchical} & 66.32 & 70.62 & & 103.90 & 175.92\\ [0.5em]
      {\em Our models}\\
      4. Attentional DCLM (\dam) & 64.31 & 68.32 & & 96.47 & {\bf 170.99}\\
      5. Context-to-output DCLM (\scmo) & 64.37 & 68.49 & & {\bf 95.10} & 173.52 \\
      6. Context-to-context DCLM (\scmh) & {\bf 62.34} & {\bf 66.42} & & 96.77 & 172.88 \\
      \bottomrule
    \end{tabular}
  }
  \caption{Perplexities of the Penn Treebank (PTB) and North American News Text (NANT) data sets.}
  \label{tab:ppl}
\end{table}

\autoref{tab:ppl} present the results on language modeling perplexity. The best perplexities are given by the context-to-context DCLM on the PTB data set (line 6 in~\autoref{tab:ppl}), and attentional DCLM on the NANT data set (line 4 in \autoref{tab:ppl}). All DCLM-based models achieve better perplexity than the prior work. While the improvements on the PTB dataset are small in an absolute sense, they consistently point to the value of including multi-level context information in language modeling. The value of context information is further verified by the model performance on the NANT dataset. Of interest is the behavior of the attentional DCLM on two data sets. This model combines both the context-to-context and context-to-output mechanisms. Theoretically, \dam is considerably more expressive then the \scmo~and \scmh. On the other hand, it is also complex to learn and innately favors large data sets.



\begin{figure}
  \centering
  \includegraphics[width=0.48\textwidth]{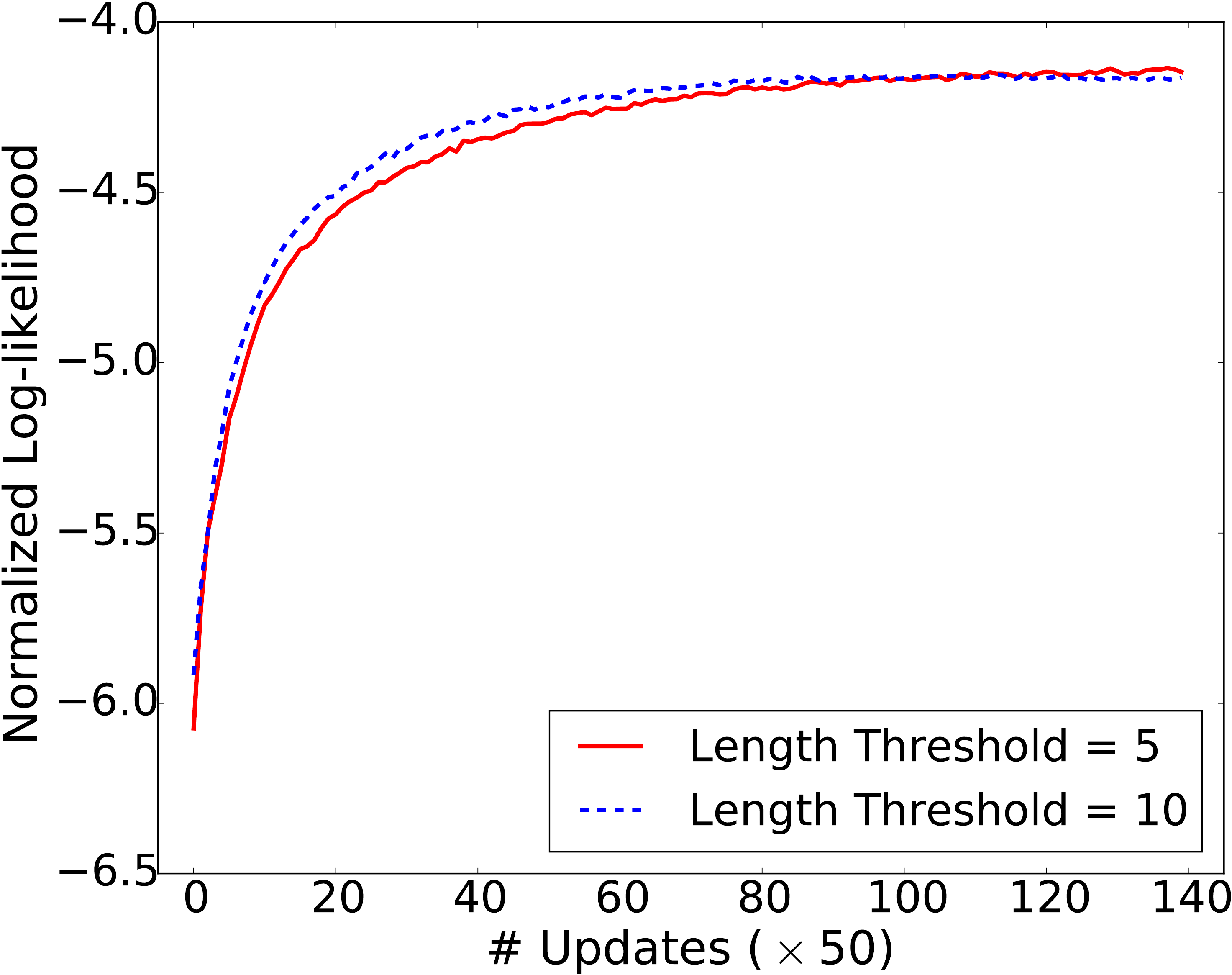}
  \caption{Effect of length thresholds on predictive log-likelihood on the PDTB development set.}
  \label{fig:diff-length}
\end{figure}

In all the results reported in \autoref{tab:ppl}, the document length threshold was fixed as $L=5$, meaning that documents were partitioned into subsequences of five sentences. We were interested to know whether our results depended on this parameter. Taking $L=1$ would be identical to the standard RNNLM, run separately on each sentence. To test the effect of increasing $L$, we also did an empirical comparison between $L=5$ and $L=10$ with \scmh. \autoref{fig:diff-length} shows the two curves on the PTB development set. The $x$-axis is the number of updates on \scmh~with the PTB training set. The $y$-axis is the mean per-token log-likelihood given by~\autoref{eq:rnnlm-emit} on the development set. As shown in this figure, $L=10$ seems to learn more quickly per iteration in the beginning, although each iteration is more time-consuming, due to the need to backpropagate over longer documents. However, after a sufficient number of updates, the final performance results are nearly identical, with a slight advantage to the $L=5$ setting. This suggests a tradeoff between the amount of contextual information and the ease of learning.

\subsection{Local Coherence Evaluation}
\label{subsec:coh-eval}
The long-term goal of coherence evaluation is to predict which texts are more coherent, and then to optimize for this criterion in multi-sentence generation tasks such as summarization and machine translation. A well-known proxy to this task is to try to automatically distinguish an original document from an alternative form in which the sentences are scrambled~\citep{barzilay2008modeling,li2014model}. Multi-sentence language models can be applied to this task directly, by determining whether the original document has a higher likelihood; no supervised training is necessary.

We adopt the specific experimental setup proposed by \citet{barzilay2008modeling}. To give a robust model comparison with the limited number of documents available in the PTB test set, we employ bootstrapping~\citep{davison1997bootstrap}. First, a new test set $\mathcal{D}^{(\ell)}$ is generated by sampling the documents from the original test set with replacement. Then, we shuffled the sentences in each document $d \in \mathcal{D}^{(\ell)}$ to get a pseudo-document $d'$. The combination of $d$ and $d'$ is a single test example. We repeated the same procedure to produce 1,000 test sets, where each test set includes 155 pairs --- one for each document in the PTB test set. Since each test instance is a pairwise choice, a random baseline will have expected accuracy of 50\%. 

To evaluate the models proposed in this paper, we use the configuration with the best development set perplexity, as shown in \autoref{tab:ppl}. The results of accuracy and standard deviation are calculated over 1,000 resampled test sets. As shown in~\autoref{tab:coherence}, the best accuracy is 83.26\% given by \scmh, which also gives the smallest standard deviation 3.77\%. 
Furthermore, all DCLM-based models significantly outperform the RNNLM with $p<0.01$ given by a two-sample one-side z-test on the bootstrap samples. 
In addition, the \scmh~and \scmo~are outperform the \hrnnlm\ with $p<0.01$ with statistic $z=36.55$ and $31.26$ respectively.

\newtext{In addition, we also evaluated the models trained on the NANT dataset on this coherence evaluation task. With the same 1,000 test sets, the best accuracy number across different models is $72.85\%$ obtained from \scmo. Compare to the results in \autoref{tab:coherence}, we believe the performance drop is due to the domain mismatch. Even though PTB and NANT are both corpora collecting from news articles, they have totally different distributions on words, sentence lengths and even document lengths as shown in \autoref{tab:dataset}.}

Unlike some prior work on coherence evaluation~\citep{li2014model,li2015hierarchical,lin2015hierarchical},
our approach is not trained on supervised data. Supervised training might therefore improve performance further. However, we emphasize that the real goal is to make automatically-generated translations and summaries more coherent, and we should therefore avoid overfitting on this artificial proxy task.


\begin{table}
  \centering
  {\footnotesize
    \begin{tabular}{lll}
      \toprule
      & \multicolumn{2}{l}{Accuracy}\\
      \cmidrule{2-3}
      Model & Mean (\%) & Standard deviation (\%) \\
      \midrule
      {\em Baselines}\\
      1. RNNLM w/o sentence boundary (\drnnlm) & 72.54 & 8.46\\
      2. Hierarchical RNNLM (\hrnnlm)~\citep{lin2015hierarchical} & 75.32 & 4.42\\ [0.5em]
      {\em Our models} \\
      3. Attentional DCLM (\dam)$^{\dagger}$ & 75.51 & 4.12\\
      4. Context-to-output DCLM (\scmo){$^{\dagger\ast}$} & 81.72 & 3.81\\
      5. Context-to-context DCLM (\scmh){$^{\dagger\ast}$} & 83.26 & 3.77\\
      \bottomrule
      $^{\dagger}$ significantly better than \drnnlm~with p-value $<0.01$\\
      $^{\ast}$ significantly better than \hrnnlm~with p-value $<0.01$
    \end{tabular}
  }
  \caption{Coherence evaluation on the PTB test set. The reported accuracies are calculated from 1,000 bootstrapping test sets (as explained in text).}
  \label{tab:coherence}
\end{table}

\section{Related Work}
\label{sec:related}

Neural language models (NLMs) learn the distributed representations of words together with the probability function of word sequences. In the NLM proposed by \citet{bengio2003neural}, a feed-forward neural network with a single hidden layer was used to calculate the language model probabilities. One limitation of this model is only fixed-length context can be used. Recurrent neural network language models (RNNLMs) avoid this problem by recurrently updating a hidden state~\citep{mikolov2010recurrent}, thus enabling them to condition on arbitrarily long histories. In this work, we make a further extension to include more context with a recurrent architecture, by allowing multiple pathways for historical information to affect the current word. A comprehensive review of recurrent neural networks language models is offered by \citet{de2015survey}.

Conventional language models, including the models with recurrent structures~\citep{mikolov2010recurrent}, limit the context scope within a sentence. This ignores potentially important information from preceding text, for example, the previous sentence. 
Targeting speech recognition, where contextual information may be especially important, \citet{mikolov2012context} introduce the \emph{topic-conditioned \rnnlm}, which incorporates a separately-trained latent Dirichlet allocation topic model to capture the broad themes of the preceding text. Our focus here is on discriminatively-trained end-to-end models.

\citet{lin2015hierarchical} recently introduced a document-level language model, called hierarchical recurrent neural network language model (\hrnnlm). As in our approach, there are two channels of information: a RNN for modeling words in a sentence, and another recurrent model for modeling sentences, based on a bag-of-words representation of each sentence. (Contemporaneously to our paper, \cite{wang2015larger} also construct a bag-of-words representation of previous sentences, which they then insert into a sentence-level LSTM.) Our modeling approach is more unified and compact, employing a single recurrent neural network architecture, but with multiple channels for information to feed forward into the prediction of each word. We also go further than this prior work by exploring an attentional architecture~\citep{bahdanau2015neural}.

Moving away from the specific problem of language modeling, we briefly consider other approaches for modeling document content. \citet{li2014model} propose to use a convolution kernel to summarize sentence-level representations for modeling a document. The model is for coherence evaluation, in which the parameters are learned via supervised training. Related convolutional architectures for document modeling are considered by \cite{denil2014modelling} and \cite{tang2015document}. Encoder-decoder architectures provide an alternative perspective, compressing all the information in a sequence into a single vector, and then attempting to decode the target information from this vector; while this idea has notably applied in machine translation~\citep{cho2014learning}, it can also be employed for coherence modeling~\citep{li2015hierarchical}. The hierarchical sequence-to-sequence model of \cite{li2015hierarchical} conditions the start word of each sentence on contextual information provided by the encoder, but does not apply this idea to language modeling. Different from the models with hierarchical structures, paragraph vector~\citep{le2014distributed} encodes a document to a numeric vector by discarding document structure and only retaining topic information. 


%



\section{Conclusion}
Contextual information beyond the sentence boundary is essential to document-level text generation and coherence evaluation. We propose a set of document-context language models (DCLMs), which provide various approaches to incorporate contextual information from preceding texts. Empirical evaluation with perplexity shows that the DCLMs give better word prediction as language models, in comparison with conventional RNNLMs; performance is also good on unsupervised coherence assessment. Future work includes testing the applicability of these models to downstream applications such as summarization and translation.

\paragraph{Acknowledgments} This work was initiated during the 2015 Jelinek Memorial Summer Workshop on Speech and Language Technologies at the University of Washington, Seattle, and was supported by Johns Hopkins University via NSF Grant No IIS 1005411, DARPA LORELEI Contract No HR0011-15-2-0027, and gifts from Google, Microsoft Research, Amazon and Mitsubishi Electric Research Laboratory. It was also supported by a Google Faculty Research award to JE.

\bibliography{ref}
\bibliographystyle{iclr2016_workshop}

\end{document}